%% file: acl_latex.tex
\documentclass[11pt]{article}

\usepackage[final]{acl}

\usepackage{times}
\usepackage{helvet}  
\usepackage{courier}  
\usepackage{latexsym}
\usepackage{natbib}
\usepackage{caption}
\usepackage{xspace}
\usepackage{algorithm}
\usepackage{amsfonts}
\usepackage{algorithmic}
\usepackage{cuted}
\usepackage{amsmath}
\usepackage{mathtools}
\usepackage{booktabs}
\usepackage{multirow}
\usepackage{graphicx}
\usepackage{xcolor}
\usepackage{newfloat}
\usepackage{listings}
\usepackage[table,dvipsnames]{xcolor}

\newcommand{\ourapproach}{\textsc{D-QReLO}\xspace}
\definecolor{mygray}{gray}{0.94}
\usepackage[T1]{fontenc}

\usepackage[utf8]{inputenc}

\usepackage{microtype}
\usepackage{fontawesome5} 
\usepackage{seqsplit}

\usepackage{inconsolata}

\usepackage{graphicx}

\definecolor{reda}{RGB}{192,0,0}

\title{\ourapproach: Training- and Data-Free Delta Compression for Large Language Models via Quantization and Residual Low-Rank Approximation}

\author{
    Junlin Li$^1$
    \quad Shuangyong Song$^2$
    \quad Guodong Du$^3$\textsuperscript{\texorpdfstring{\faIcon[regular]{envelope}}{}} 
    \quad Ngai Wong$^4$\\
    \quad \textbf{Xuebo Liu}$^1$
    \quad \textbf{Yongxiang Li}$^2$
    \quad \textbf{Min Zhang}$^1$
    \quad  \textbf{Jing Li}$^1$\textsuperscript{\texorpdfstring{\faIcon[regular]{envelope}}{}} 
    \quad  \textbf{Xuelong Li}$^2$\textsuperscript{\texorpdfstring{\faIcon[regular]{envelope}}{}} 
    \\$^{1}$Harbin Institute of Technology, Shenzhen, China \quad
    $^{2}$TeleAI of China Telecom    
    \\$^{3}$The Hong Kong Polytechnic University \quad
     $^{4}$The University of Hong Kong \\
    \texttt{leejunlin27@gmail.com} \quad \texttt{jingli.phd@hotmail.com}  
}

\begin{document}
\maketitle
\begin{abstract}
Supervised Fine-Tuning (SFT) accelerates task-specific large language models (LLMs) development, but the resulting proliferation of fine-tuned models incurs substantial memory overhead.
Delta compression addresses this by retaining a single pre-trained LLM with multiple compressed delta weights. 
However, existing methods fail on models fine-tuned with large-scale datasets.
We find that larger SFT data scale amplifies delta parameter magnitude, singular values, and entropy, exacerbating compression errors.
To tackle this, we propose \ourapproach (\textbf{D}elta Compression via  \textbf{Q}uantization and \textbf{Re}sidual \textbf{Lo}w-Rank), a novel training- and data-free delta compression method.
It combines coarse-grained one-bit quantization to capture the dominant structure of the delta, followed by compensated residual low-rank approximation to recover fine-grained details from the smaller residual error.
Experiments on various LLMs spanning dense and MoE architectures across multiple domains under this challenging setting demonstrate that \ourapproach outperforms existing methods.
Moreover, we establish key design principles for delta compression through extensive empirical analysis, demonstrating how task difficulty, architecture, and layer positioning create predictable patterns that can guide optimal compression strategies in production systems.

\let\thefootnote\relax\footnotetext{\faIcon[regular]{envelope}~Corresponding authors.}
\end{abstract}

\section{Introduction}
Large Language Models (LLMs)~\cite{touvron2023llama,achiam2023gpt,du2025nps} have recently become a cornerstone in AI, fundamentally reshaping how various downstream tasks are approached. Leveraging fine-tuning, pre-trained LLMs now yield diverse specialized models~\cite{hui2024qwen2,DBLP:conf/acl/LeeWLZ24,du2025graftllm}. This paradigm significantly lowers model customization barriers, accelerating AI deployment across industries. 
However, this flexibility paradigm introduces significant engineering challenges. 
Real-world applications often demand simultaneously storing and deploying numerous specialized models for complex needs~\cite{DBLP:conf/acl/LeeWLZ25,farr2024llm,DBLP:conf/acl/LeeWLZ24}.
For example, it is essential to store every iterative model version to facilitate rollbacks, as well as to cooperatively deploy multiple specialized models in multi-stage reasoning or multi-agent collaboration tasks~\cite{DBLP:conf/iclr/ChanCSYXZF024}.

     \begin{figure}[t]
		\centering
		\includegraphics[width=1\columnwidth]{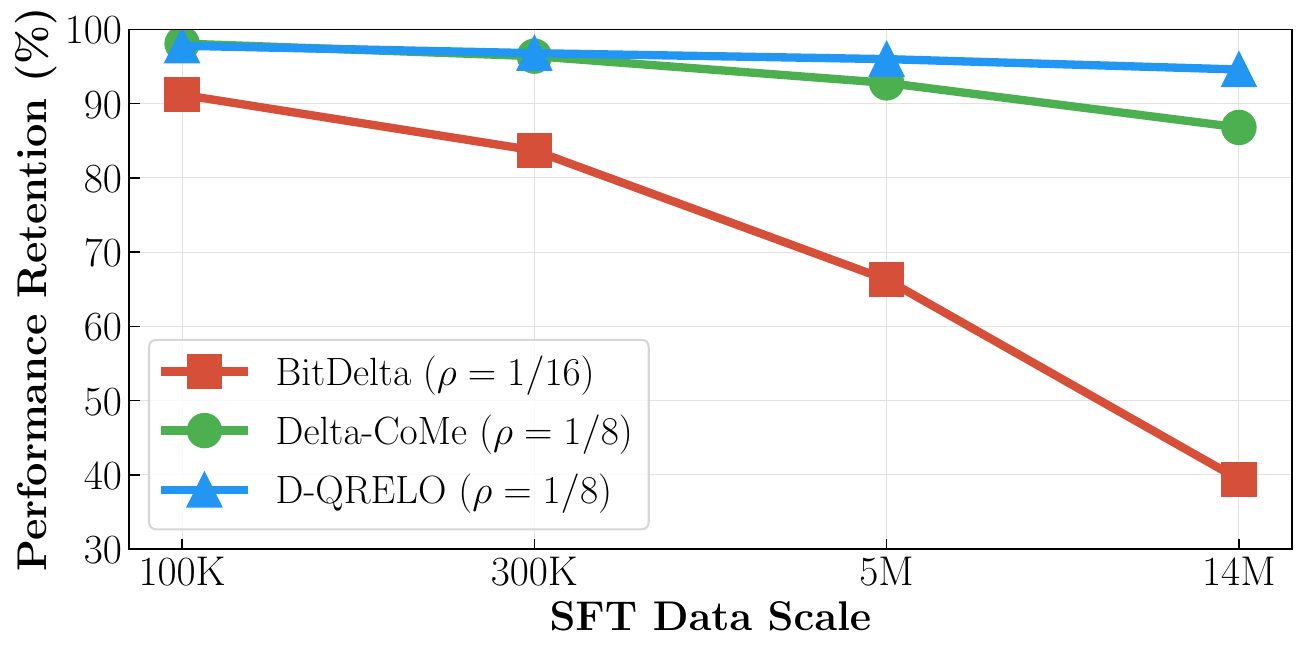}

		\caption{   Impact of SFT data scale on delta compression for the same base LLM. $\rho$ is the compression ratio.
		}
		\label{fig:motivation}
	\end{figure}
    \input{table/motivation}

Recently, delta compression~\cite{DBLP:conf/eurosys/YaoHK25,yang2025impart} has emerged to tackle these challenges by exploiting redundancy in delta parameters—the differences between fine-tuned model and pre-trained model weights. 
Its goal is to store only a single pre-trained model alongside multiple compressed delta parameter sets, thereby reducing the memory footprint for storing and deploying numerous fine-tuned models.
For instance, BitDelta~\cite{DBLP:conf/nips/LiuXLL0DC24} compresses delta parameters to one-bit quantization. Delta-CoMe~\cite{DBLP:conf/nips/PingWWHXY0CL024} further improves compression efficiency by leveraging mixed-precision SVD of delta parameters.

However, these methods show a significant performance drop when applied to powerful SFT models (e.g., Qwen3, DeepSeek-R1-Distill). To understand this, we analyze these models and find their delta parameters magnitudes are consistently larger (mean absolute values: ~0.002-0.004) compared to the LLMs these methods originally evaluated (mostly $<$0.0008). This highlights an acknowledged but critically underappreciated issue~\cite{DARE}: the magnitude of the delta parameters profoundly impacts compression performance.

To delve deeper into this phenomenon, we conduct an extensive exploration: under the same experimental setup, we fine-tune LLaMA 3.1 8B with OpenMathInstruct-2~\cite{DBLP:conf/iclr/ToshniwalDMKAG25} datasets of 100K, 300K, 5M, and 14M samples, respectively. Table~\ref{tab:motivation} presents the magnitude of delta parameters, their singular values and entropy. Figure~\ref{fig:motivation} shows the performance of various methods on these models. Key observations follow:
\vspace{-0.2mm}
\begin{itemize}
    	\item Under same setting, more fine-tuning data leads to larger magnitude of delta parameters, their singular values, and their entropy, indicating that the delta parameters are becoming increasingly complex and information-rich.
        \vspace{-0.2mm}
    	\item As these delta properties intensify, the performance of prior methods declines significantly.
\end{itemize} 
\vspace{-0.1mm}

This performance degradation stems from the inherent limitations of quantization and SVD, which incur larger errors when handling delta parameters with greater numerical scales and complexity.

Given these critical insights, we propose \ourapproach, a delta compression method combining quantization and compensated residual low-rank approximation. 
\ourapproach begins with coarse-grained bit quantization to capture the dominant signs and initial magnitude information of the delta parameters. This step serves as a fast, adaptive encoding mechanism that efficiently handles large-magnitude values. 
Subsequently, we observe that the residual error exhibits a substantially narrower numerical range than the original delta parameters, rendering it well-suited for efficient low-rank approximation via SVD.
By applying SVD to this residual, we can precisely reconstruct the fine-grained details lost during the initial quantization.
Crucially, unlike prior methods, \ourapproach achieves both training- and data-free compression, which enhances its generality and efficiency.

Extensive experiments on powerful SFT models spanning dense and MoE architectures trained with large-scale data show that \ourapproach consistently outperforms prior methods at the same compression ratio, advancing the Pareto frontier of delta compression and demonstrating superior efficiency-performance trade-offs.
These models cover tasks in reasoning, alignment, domain knowledge, and multimodal understanding, highlighting  \ourapproach's strong generalization. 
Evaluations show that \ourapproach yields multi-fold savings in GPU memory, along with significant inference speedup.
Furthermore, to explore key design principles for delta compression, we conduct a comprehensive empirical study grounded in \ourapproach.

The main contributions of this paper include:
\vspace{-\topsep} 

    \begin{itemize}
        \setlength{\itemsep}{0pt}   
    \setlength{\parsep}{0pt}    
    \setlength{\partopsep}{0pt} 
    \item We provide insights into the impact of SFT data scale on delta compression efficiency.
    \item We propose a novel delta compression method, \ourapproach, that combines quantization and residual low-rank approximation.
    \item We conduct the first comprehensive empirical study of delta compression.
    \end{itemize} 
    
\section{Related Work}
\paragraph{Delta-Compression.}
Delta compression methods are typically categorized into pruning, quantization, and low-rank approximation. Pruning methods~\cite{PCB,liu2024me,DBLP:conf/acl/Li0LGWWLTAHZ25} like DARE~\cite{DARE} reduce model size by eliminating redundant delta parameters while maintaining performance. Quantization approaches, known for their efficiency and hardware friendliness, compress delta parameters into low-bit formats.
GPT-Zip~\cite{isik2023gpt} pioneers 2-bit quantization paired with sparsification to compress delta weights while maintaining performance. BitDelta~\cite{DBLP:conf/nips/LiuXLL0DC24} pushes compression further by quantizing delta parameters down to one-bit, and refines scale factors through knowledge distillation. 
Low-rank approximation methods~\cite{ryu2023efficient} like Delta-CoMe~\cite{DBLP:conf/nips/PingWWHXY0CL024} employ mixed-precision quantization that adapts bit-width allocation according to the singular value spectrum of delta matrices.
However, large-scale fine-tuning on massive datasets often amplifies the magnitude of delta parameters, breaking prior methods' assumptions about their value range and error distribution—resulting in suboptimal compression.
To address this, we propose \ourapproach, designed to robustly handle the expanded numerical range of delta parameters in powerful SFT models.

\paragraph{Quantization and Low Rank Approximation.}
Quantization~\cite{DBLP:conf/mlsys/0002TTYCWXDG024} and low-rank~\cite{DBLP:conf/icml/LiYZLHCZ23} approximation are complementary techniques that are often combined to enhance the compression of LLM. 
For example, CALDERA~\cite{DBLP:conf/nips/SahaSSGP24} decomposes a weight matrix as $W \approx Q + LR$, where $Q$ is a quantized full-rank matrix and $LR$ is a low-precision low-rank term, optimized alternately by quantizing $W - LR$ and factorizing $W - Q$. LQ-LoRA~\cite{DBLP:conf/iclr/GuoGXK24} adopts a low-rank plus quantized matrix decomposition, where the quantized component is kept fixed during fine-tuning while only the low-rank component is updated. 
Additionally, LQER~\cite{DBLP:conf/icml/ZhangCCZ24} introduces an activation-induced scale matrix to optimize the singular value distribution of the quantization error, effectively restoring LLM performance.
In this work, we pioneer the first method to combine quantization with residual low-rank approximation, specifically tailored for delta parameters compression.

\begin{figure*}[htbp] 
    \centering
    \includegraphics[width=0.99\textwidth]{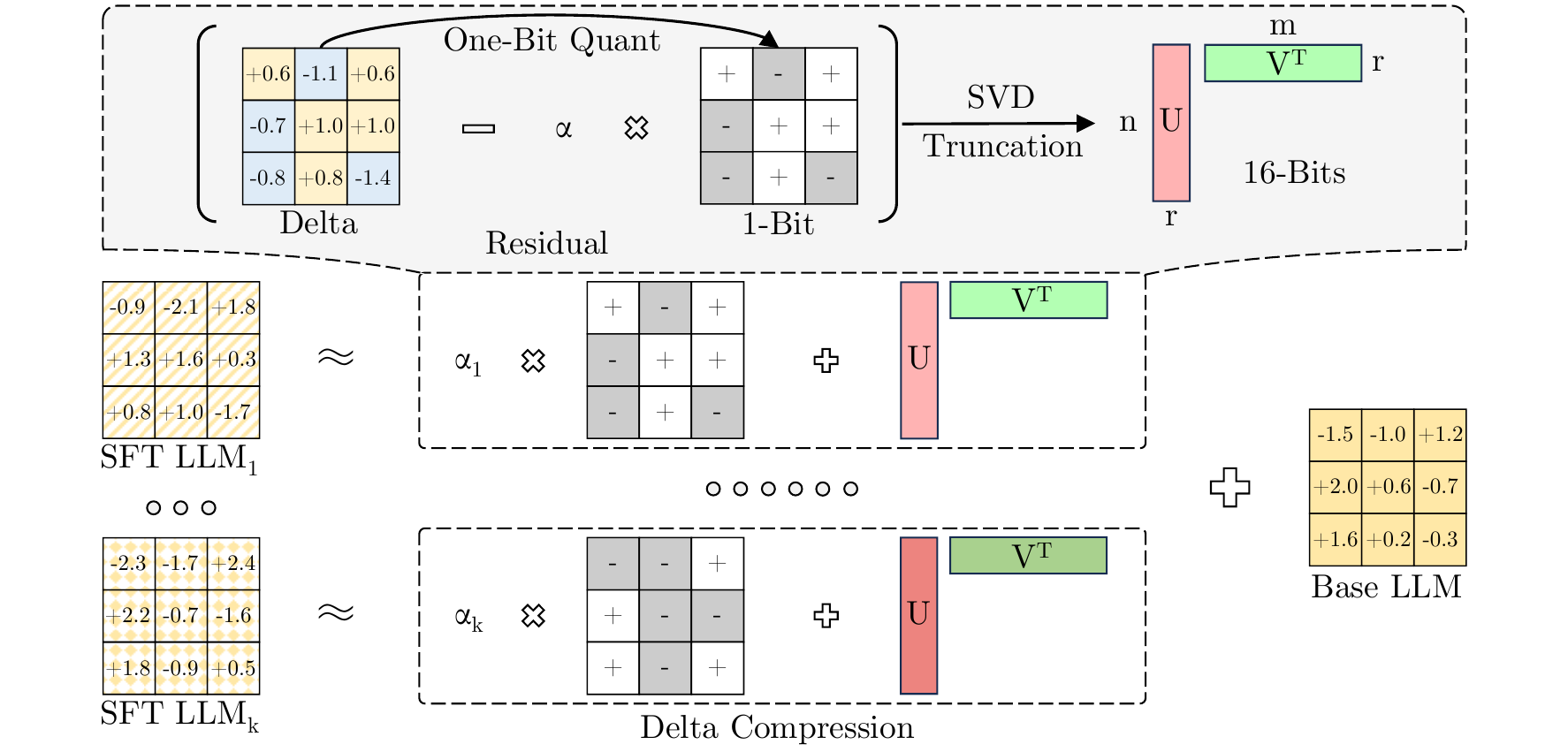} 

    \caption{Overview of \ourapproach. \ourapproach involves an initial one-bit quantization of the delta parameters, followed by SVD truncation applied to the resulting residuals to derive a low-rank matrix.
    }

    \label{fig:method}
\end{figure*}

\section{Methodology}
\subsection{Preliminaries}
Given a scenario in which a pre-trained LLM $\phi_{Pre}$ is independently fine-tuned for a set of downstream tasks $\{ \mathcal{T}_i \}_{i=1}^K$. 
This yields a collection of fine-tuned models $\{ \phi^{(i)} \}_{i=1}^K$, each customized for a specific task. 
Directly storing all $K$ full-precision models incurs a storage cost of $K \times \mathcal{M}$, where $\mathcal{M}$ denotes the number of parameters in the full model. 
To reduce this overhead, we first represent each fine-tuned LLM as the combination of the pre-trained LLM $\phi_{Pre}$ and its corresponding delta parameters $\Delta$, which are defined as:
\begin{equation}
    \Delta^{(k)} = \phi^{(k)} - \phi_{Pre}, \quad k = 1, 2, \dots, K,
\end{equation}
where delta parameters $\Delta^{(k)}$ captures the task-specific deviation from the pre-trained LLM.

We then apply a delta compression function $\mathcal{C}(\cdot; \rho)$ to transform each $\Delta^{(k)}$ into compressed delta parameters $\hat{\Delta}^{(k)}$:
\begin{equation}
    \hat{\Delta}^{(k)} = \mathcal{C}(\Delta^{(k)}; \rho),
\end{equation}
where $\rho$ denotes the compression ratio. The decompression procedure reconstructs an approximate fine-tuned LLM as:
\begin{equation}
    \hat{\phi}^{(k)} = \phi_{Pre} + \hat{\Delta}^{(k)}.
\end{equation}

By storing a single pre-trained LLM $\phi_{Pre}$ with a set of compressed delta parameters $\{ \hat{\Delta}^{(1)}, \dots, \hat{\Delta}^{(K)} \}$, this framework reduces total storage to $(1 + \rho K) \times \mathcal{M}$.

\subsection{Motivation}
As shown in Figure~\ref{fig:motivation} and Table~\ref{tab:motivation}, the performance of prior methods degrades as the magnitude of delta parameters and their singular values increases, due to their struggle in maintaining sufficient accuracy on larger numerical scales.

For a delta parameters ($\Delta_1$) approximated by its top-$k$ singular values ($(\Delta_1)_k$), the Frobenius norm of the reconstruction error is defined as:
\begin{equation}
 \|\Delta_1 - (\Delta_1)_k\|_F^2 = \sum_{i=k+1}^{\text{rank}(\Delta_1)} \sigma_{1,i}^2 
\end{equation}

If a delta parameters $\Delta_2$ possesses singular values $\sigma_{2,i}$ consistently larger than $\Delta_1$'s (e.g., $\sigma_{2,i} = c \cdot \sigma_{1,i}$ for $c>1$), then, even with the same $k$ singular values retained, its absolute reconstruction error will be significantly larger:
\begin{equation}
\scalebox{0.922}{
$ 
\begin{split}
&\|\Delta_2 \!-\! (\Delta_2)_k\|_F^2 \!=\! \smashoperator{\sum_{i=k+1}^{\text{rank}(\Delta_2)}} \sigma_{2,i}^2 \!=\! \smashoperator{\sum_{i=k+1}^{\text{rank}(\Delta_2)}} (c \!\cdot\! \sigma_{1,i})^2 \\
&= c^2 \!\! \smashoperator{\sum_{i=k+1}^{\text{rank}(\Delta_1)}} \sigma_{1,i}^2 \!=\! c^2 \|\Delta_1 \!-\! (\Delta_1)_k\|_F^2
\end{split}
$ 
} 
\end{equation}

Thus, SVD-based methods incur a greater absolute error on larger-magnitude delta parameters for a given compression ratio.
Similarly, for the one-bit quantization method BitDelta, the absolute quantization error scales proportionally with the magnitude of the original values. 
For an element $x$ quantized to $x_q$, the absolute error is $|x-x_q|$. Consequently, mapping a larger magnitude x to fixed discrete points (e.g., -1 or 1) results in a proportionally larger absolute deviation.

In summary, neither method mitigates magnitude related error accumulation effectively. To resolve this limitation, we introduce \ourapproach, a novel two-stage residual-calibrated framework targeting the core cause of performance degradation.

\subsection{\ourapproach}
Our method \ourapproach is illustrated in Figure~\ref{fig:method}. Given the delta parameters \(\Delta \in \mathbb{R}^{n \times m}\), \ourapproach's goal is to efficiently compress \(\Delta\) with a target compression ratio \(\rho\). 

\ourapproach first applies a coarse-grained one-bit quantization to \(\Delta\) to capture essential directional properties and initial magnitude scale:
\begin{equation}
\hat{\Delta}_{\text{quant}} = \alpha \odot \operatorname{Sign}(\Delta),
\end{equation}
where the element-wise sign function is defined as:
\begin{equation}
\operatorname{Sign}(\Delta_{ij}) =
\begin{cases}
+1, & \Delta_{ij} > 0, \\
-1, & \Delta_{ij} \leq 0,
\end{cases}
\end{equation}
and the scaling factor \(\alpha\) is determined as the mean absolute value of \(\Delta\):
\begin{equation}
\alpha = \frac{1}{nm} \sum_{i=1}^n \sum_{j=1}^m |\Delta_{ij}|,
\end{equation}
This choice of \(\alpha\) minimizes the Frobenius-norm quantization error between \(\Delta\) and its one-bit quantized approximation $\hat{\Delta}_{\text{quant}}$:
\begin{equation}
\alpha = \arg\min_{\alpha} \| \Delta - \alpha \cdot \operatorname{Sign}(\Delta) \|_F^2.
\end{equation}

\input{table/main}
The remaining fine-grained information and the error introduced by coarse quantization are preserved in the residual matrix \(R\):
\begin{equation}
R = \Delta - \hat{\Delta}_{\text{quant}}.
\end{equation}
Thanks to the effective initial quantization, \(R\)'s numerical range is significantly smaller than that of the original \(\Delta\), making it suitable for efficient low-rank approximation.

\ourapproach then computes the SVD of residual matrix \(R\):
\begin{equation}
R = U \Sigma V^\top,
\end{equation}
where \(U \in \mathbb{R}^{n \times n}\), \(\Sigma \in \mathbb{R}^{n \times m}\) is diagonal with singular values \(\sigma_{R,1} \geq \sigma_{R,2} \geq \cdots\), and \(V \in \mathbb{R}^{m \times m}\).

To reduce dimensionality, \ourapproach truncates the SVD of \(R\) to rank
\begin{equation}
r = \left\lceil \frac{n \cdot m \cdot \rho_1}{n + m} \right\rceil,
\end{equation}
choosing $r$ so that the low-rank parameters $r(n+m) + r$ meet the target compression ratio $\rho_1$ of the original $nm$ parameters, the additional $r$ parameters are negligible for large $n,m$.

\ourapproach then extracts the truncated matrices:
\begin{equation}
U_r = U[:, :r], \quad \Sigma_r = \Sigma[:r, :r], \quad V_r = V[:, :r].
\end{equation}
This yields a rank-$r$ approximation of the residual \(R\) as:
\begin{equation}
\hat{R}_{\text{lowrank}} = U_r \Sigma_r (V_r)^\top.
\end{equation}

The final compressed delta parameters \(\hat{\Delta}\) combines the quantization and residual low-rank approximation:
\begin{equation}
\hat{\Delta} = \hat{\Delta}_{\text{quant}} + \hat{R}_{\text{lowrank}} .
\end{equation}

The overall compression ratio $\rho$ combines the low-rank component $\rho_1$ and the one-bit quantization contribution:
\begin{equation}
\rho = \rho_1 + \frac{1}{b}
\end{equation}
where $b$ denotes the number of bits per full-precision weight (e.g., $b=16$ for FP16/BF16 precision).
For vector parameters, we select parameters for compression by magnitude at a target ratio of $\rho$.

\section{Experimental Setup}
\paragraph{Baselines.}
We evaluate several baselines: pruning methods (random, magnitude, Wanda~\cite{DBLP:conf/iclr/Sun0BK24}), SVD, BitDelta~\cite{DBLP:conf/nips/LiuXLL0DC24}, and Delta-CoMe~\cite{DBLP:conf/nips/PingWWHXY0CL024}. All methods are evaluated at a 1/8 compression ratio, as \ourapproach's design requires a ratio greater than 1/16. BitDelta is the only exception, evaluated at 1/16 due to its 1-bit quantization under FP16/BF16.
\paragraph{Tasks.}
We evaluate a set of tasks and datasets spanning various domains and levels of complexity: 
\vspace{-1.9mm}
\vspace{-\topsep} 
    \begin{itemize}
        \setlength{\itemsep}{0pt}   
    \setlength{\parsep}{0pt}    
    \setlength{\partopsep}{0pt} 
    \item Math: GSM8K~\cite{cobbe2021training}, MATH~\cite{DBLP:conf/nips/HendrycksBKABTS21} and AIME2024 span tasks from basic arithmetic to advanced competition-level problems.
    \item Code:  
    LiveCodeBench v6~\cite{livecodebench}, HumanEval~\cite{chen2021evaluating} evaluate natural language to executable code generation.
    \item Knowledge: GPQA-Diamond~\cite{rein2024gpqa}, MMLU-Pro~\cite{MMLU-Pro} assess multi-disciplinary knowledge.
    \item Alignment: IFEval~\cite{ifeval} evaluates instruction-following capability.
    \item Multi-Modal Chat: MME~\cite{fu2024mmecomprehensiveevaluationbenchmark} and TextVQA~\cite{textvqa}, evaluating joint visual-text understanding and reasoning.
    
\end{itemize}
\vspace{-3mm}
\input{table/ab1}
\paragraph{Models.}
We evaluate eight models across three tuning paradigms (standard/multimodal SFT, reasoning distillation) and two architecture types (dense, MoE). All fine-tuned on large-scale datasets and exhibit large delta parameters (average absolute value: 0.0024), details are in Table~\ref{tab:models}.

\section{Experimental Results}
\subsection{Main Results}
Our results in Table~\ref{tab:main} showcase the delta compression performance of all methods on five powerful SFT LLMs. Three additional models’ results are included in Appendix~\ref{sec:result}.
\ourapproach almost consistently shows superior performance retention across all evaluated tasks and models, achieving a 7.83\% gain over the current SOTA method Delta-CoMe.

Firstly, we observe that for delta parameters with large magnitudes, pruning (Random, Magnitude, Wanda) and low-rank (SVD) methods yield substantial performance drops (35.62\%–86.12\%).

Secondly, while BitDelta and Delta-CoMe achieve near-lossless performance in their original experiments, their efficacy significantly degrades when applied to these extensively trained SFT models. 
In contrast, \ourapproach markedly improves upon both, retaining about 94\% of the original performance. This highlights how its two-stage approach effectively mitigates the absolute error accumulation problem inherent in other methods.

Finally, \ourapproach’s outstanding performance across diverse tasks and LLM architectures underscores its superior generalization ability. This confirms the robustness of its magnitude and residual processing approach for various fine-tuned models.

\subsection{Ablation Study}
Table~\ref{tab:ab1} validates the effectiveness and strategic ordering of \ourapproach's two-stage compression strategy. 
While standalone SVD or one-bit quantization individually lead to significant performance degradation on large-magnitude delta parameters, their combination dramatically mitigates this decline. 
\begin{figure}[t]
		\centering
		\includegraphics[width=1\columnwidth]{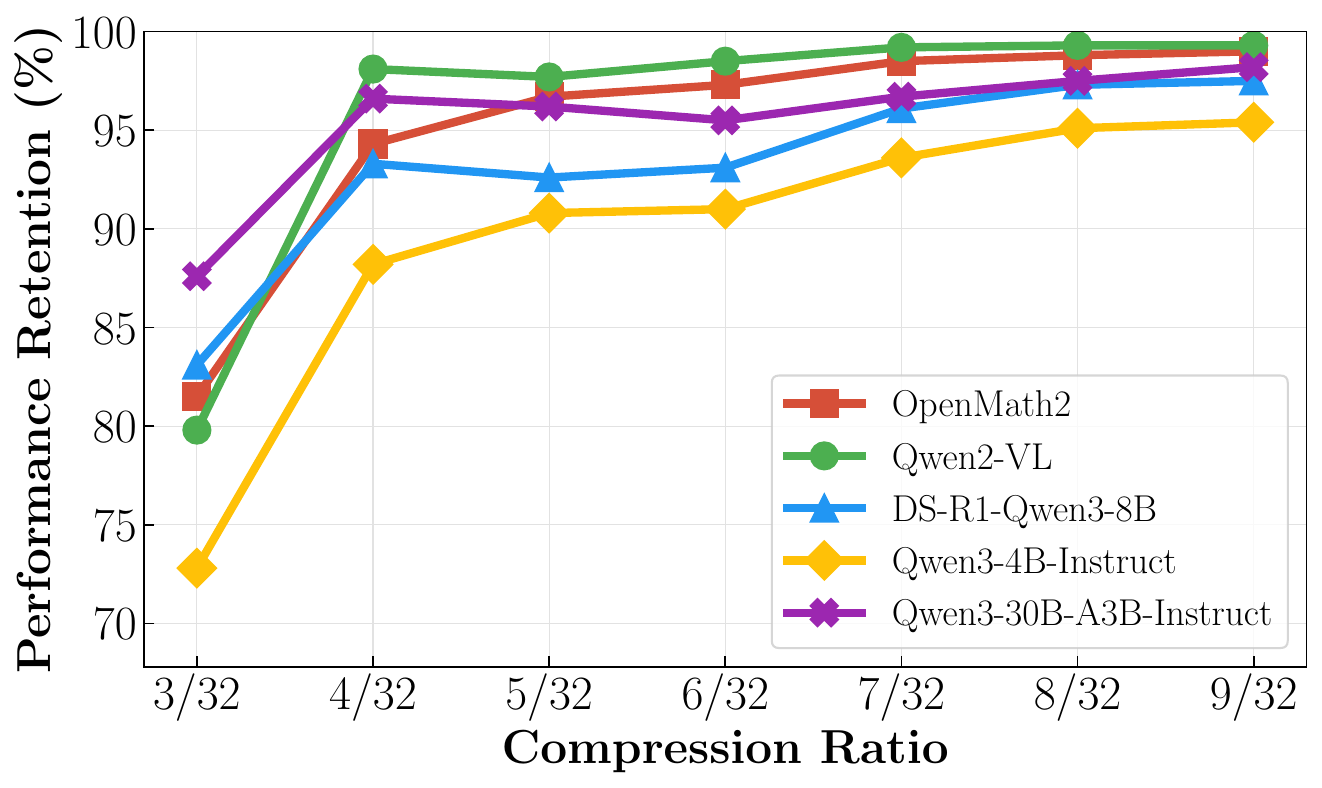}

		\caption{   Performance retention of \ourapproach across models with varying residual SVD compression ratios.
		}
        \vspace{-2mm}

		\label{fig:ab2}
	\end{figure}
Moreover, SVD proves more effective than quantization for compressing the residual. 
This is because initial one-bit quantization effectively reduces the original delta parameters' numerical scale, yielding a smaller-magnitude residual optimally suited for SVD to capture its remaining structured information. 
Conversely, SVD's residual retains high-frequency components that one-bit quantization struggles to compress accurately.

Figure~\ref{fig:ab2} shows \ourapproach's performance retention as the compression ratio varies, with one-bit quantization held constant.
We observe that performance retention generally improves with the higher compression ratio, often plateauing at higher ratios.
However, this trend is not strictly monotonic. For example, on Qwen3-30B-A3B-Instruct, increasing the ratio from 4/32 to 6/32 causes a performance drop.
This is because a higher compression ratio introduces more residual information, which may capture instability and temporarily impact performance.
This suggests that information recovery is nonlinear and influenced by residual complexity.

\subsection{Inference Speed and Memory Cost}
\input{table/memory}
Table~\ref{tab:memory_latency} compares the GPU memory and decoding latency for multiple LLaMA 3.1-8B alignment models (BF16). On a single 80G GPU, the baseline deployment lacks the capacity for even 8 models. In contrast, \ourapproach (1/8 compression ratio) drastically reduces memory overhead and significantly lowers decoding latency. Moreover, the more models deployed, the more significant the effectiveness of \ourapproach.

\section{Delta-Compression Analysis}

\subsection{Impacts of Task Difficulty}
\begin{figure}[t]
		\centering
		\includegraphics[width=1\columnwidth]{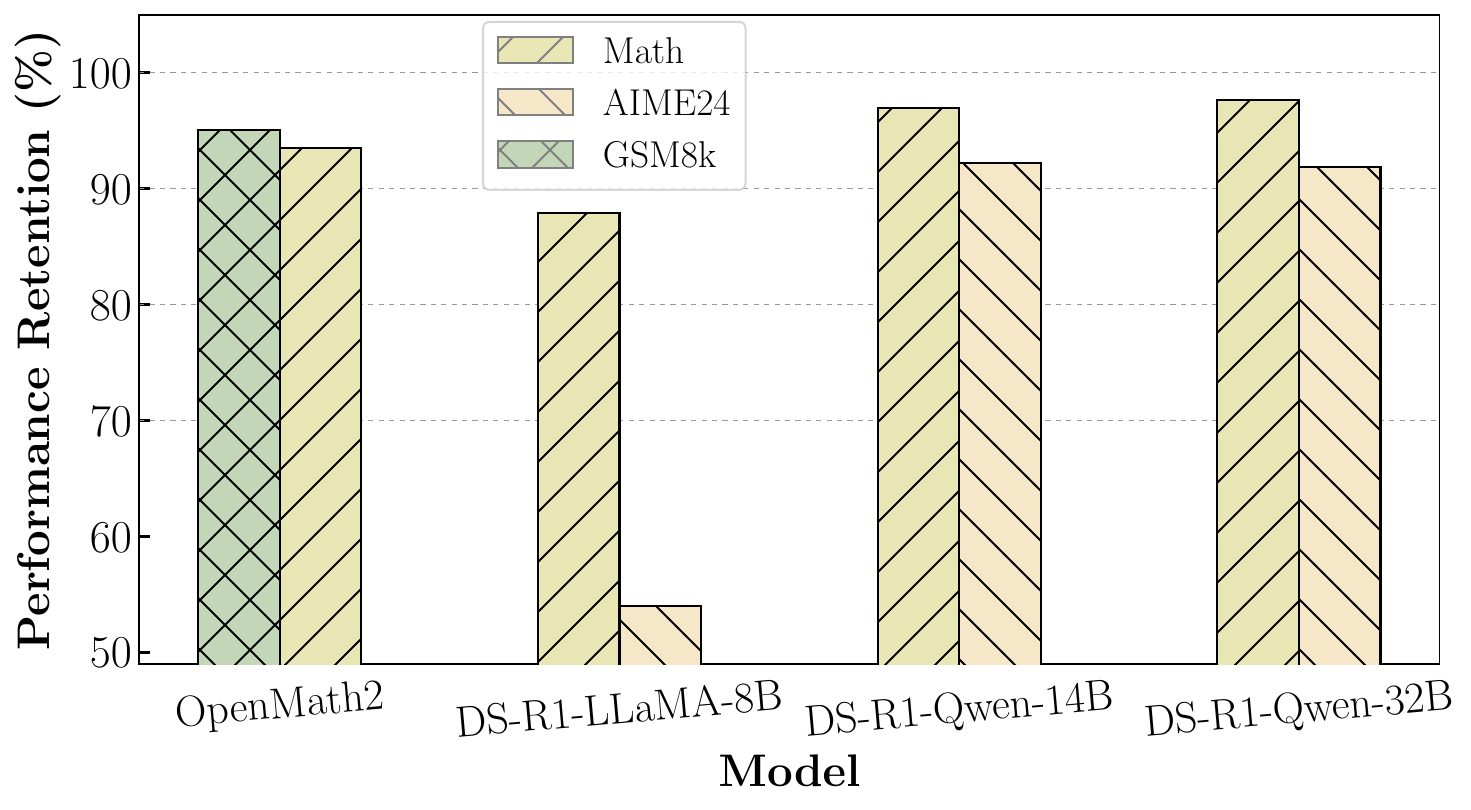}

		\caption{   Performance retention of \ourapproach on mathematical reasoning tasks of varying difficulty.
		}

		\label{fig:task}
	\end{figure}

\begin{figure}[t]
		\centering
		\includegraphics[width=1\columnwidth]{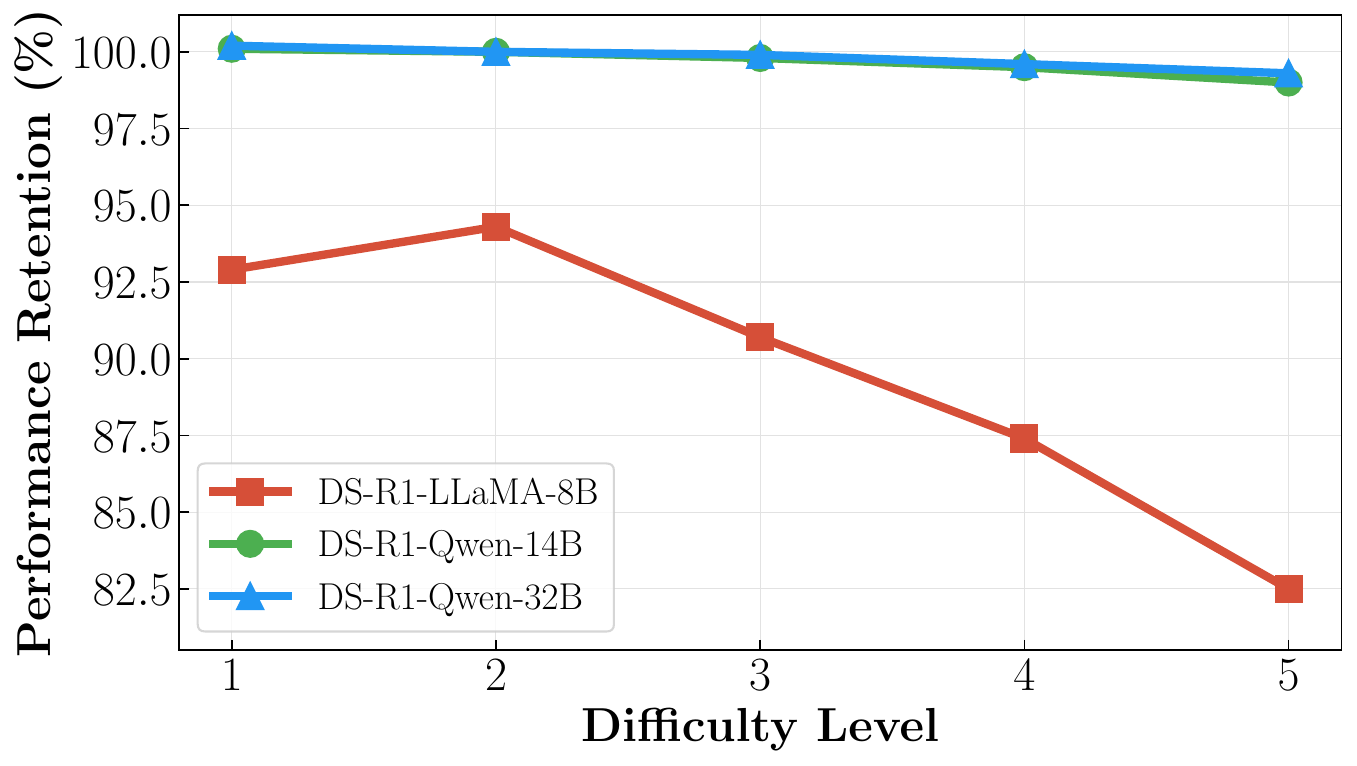}

		\caption{    Performance retention of \ourapproach on different difficulty levels on MATH.
		}

		\label{fig:math_level}
	\end{figure}

Our results, detailed in Figure~\ref{fig:task} and Figure~\ref{fig:math_level}, confirm that task difficulty directly impacts delta compression performance, with more challenging tasks consistently leading to greater performance drops across distinct mathematical benchmarks.
For instance, OpenMath2 sees a slight decline from GSM8K to MATH, while DS-R1-LLaMA-8B's performance significantly decreases from MATH to AIME24. 
This trend is further supported by a fine-grained analysis within MATH: increasing difficulty levels (1 to 5) generally correlate with reduced performance retention, most notably for DS-R1-LLaMA-8B. 
Although DS-R1-Qwen remain remarkably robust, the overall pattern highlights that the subtle and complex information required for harder tasks is more susceptible to delta compression's approximation errors.

\subsection{Impacts of the Base Pre-trained LLMs}
    
 \begin{figure}[t]
		\centering
		\includegraphics[width=1\columnwidth]{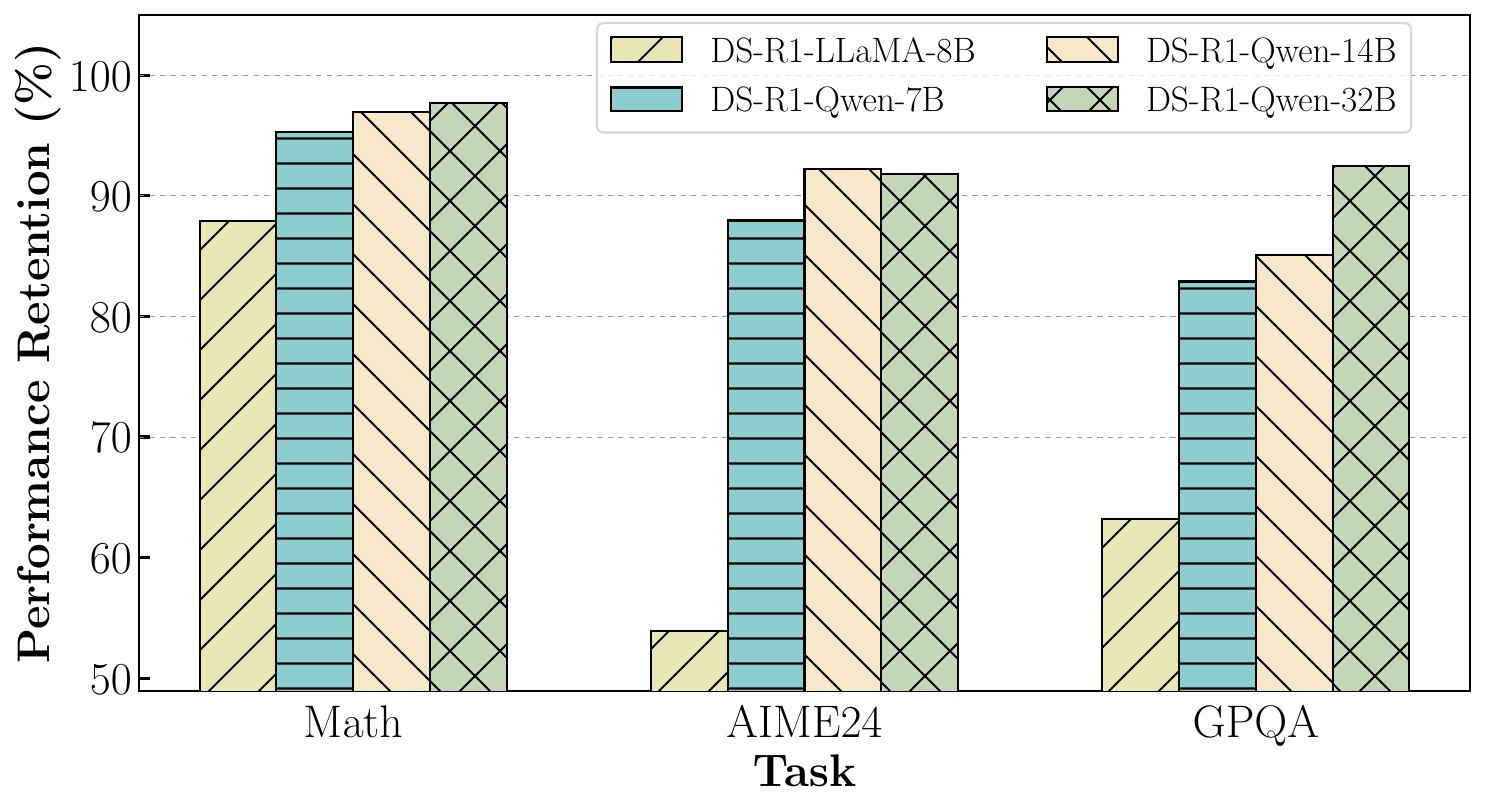}

		\caption{   Performance retention of \ourapproach across diverse base LLM architectures and sizes.
		}

		\label{fig:model_size}
	\end{figure}

Figure~\ref{fig:model_size} shows how the pre-trained model's origin and scale impact delta compression performance. 
A stark contrast emerges between LLM families, as Qwen consistently outperforms LLaMA in performance retention across all tasks.
This suggests that the Qwen base architecture, fine-tuning settings, or its pre-training characteristics, yield delta parameters inherently more structured and robust to \ourapproach's compression.
Moreover, within a family, model scale generally benefits from compressibility.
Comparing DS-R1-Qwen variants (7B–32B), larger models generally maintain slightly better performance retention, especially on GPQA. 
Hence, larger base models, with their increased capacity, more robust internal representations, and potentially greater parameter redundancy, may learn delta parameters more amenable to compression.

\subsection{Delta-Compression vs. Delta-Tuning}
\input{table/lora}
Delta compression is closely related to delta tuning, like LoRA~\cite{DBLP:conf/iclr/HuSWALWWC22}. While delta tuning aims to reduce LLM fine-tuning costs, delta compression focuses on lowering storage and inference costs for multi-model serving. 
We compare their performance by fine-tuning a LoRA model on the same dataset as OpenMath2, detailed configuration of LoRA refer to appendix~\ref{sec:lora_setting}. Table~\ref{tab:lora} shows that SVD delta compression, despite mirroring LoRA's architecture and parameter size, performs significantly worse than the LoRA model. 
In contrast, \ourapproach notably outperforms the LoRA model, closely matching the SFT model's performance. 
This suggests that for building high-performance, compact models, fully fine-tuning and then applying a well-designed delta compression method is superior to solely relying on delta tuning.

\subsection{Layer-wise Analysis}

     \begin{figure}[t]
		\centering
		\includegraphics[width=1\columnwidth]{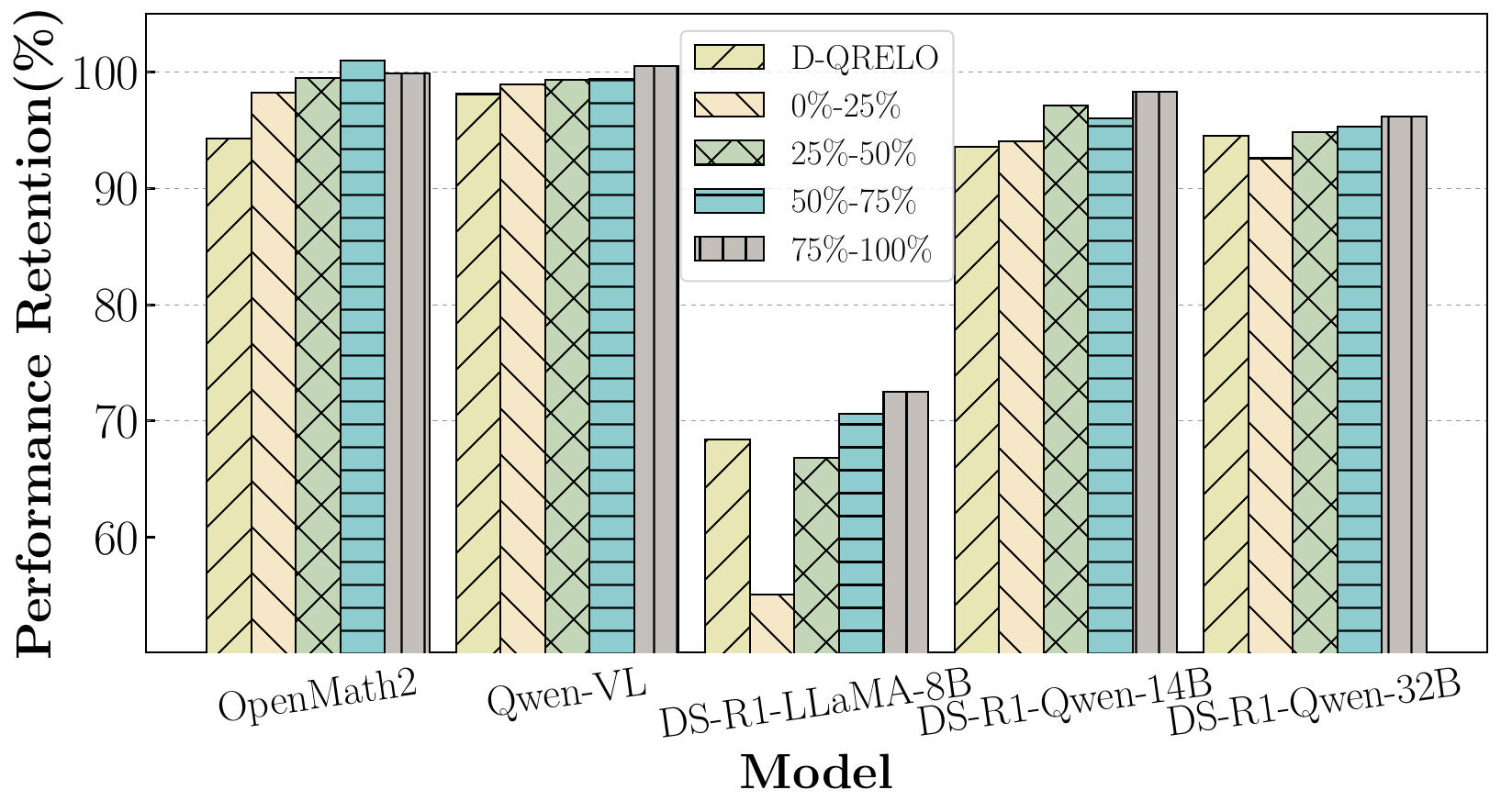}

		\caption{   Layer sensitivity of delta compression. xx\%-yy\% denotes that only the delta parameters in the first xx\% to yy\% of the model's layers are compressed.
		}
        \vspace{-2.2mm}
		\label{fig:layer}
	\end{figure}

To understand the sensitivity and robustness of different layers to delta compression, we perform an ablation study by applying \ourapproach to specific quarters of the model's layers (0\%-25\%, 25\%-50\%, 50\%-75\%, 75\%-100\%), leaving the rest uncompressed. Figure~\ref{fig:layer} highlights clear patterns in layer importance and compressibility.

Firstly, early layers are more sensitive to compression. For most models, compressing early layers causes a noticeable performance drop compared to compressing later segments.
For instance, DS-R1-LLaMA-8B suffers a significant performance drop when only its initial layers are compressed.
This suggests that delta parameters in the initial layers encode more critical, less redundant information, making them essential for preserving overall model performance and revealing a potential information bottleneck concentrated in early layers.

Secondly, later layers exhibit greater robustness and redundancy.
Compressing only these layers often achieves remarkably high performance retention, at times even surpassing that of full compression.
This suggests that delta parameters in deeper layers may exhibit greater redundancy or adapt more compressibly, with minimal performance loss—highlighting their inherent robustness.

\subsection{Module-wise Analysis}
\begin{figure}[t]
		\centering
		\includegraphics[width=1\columnwidth]{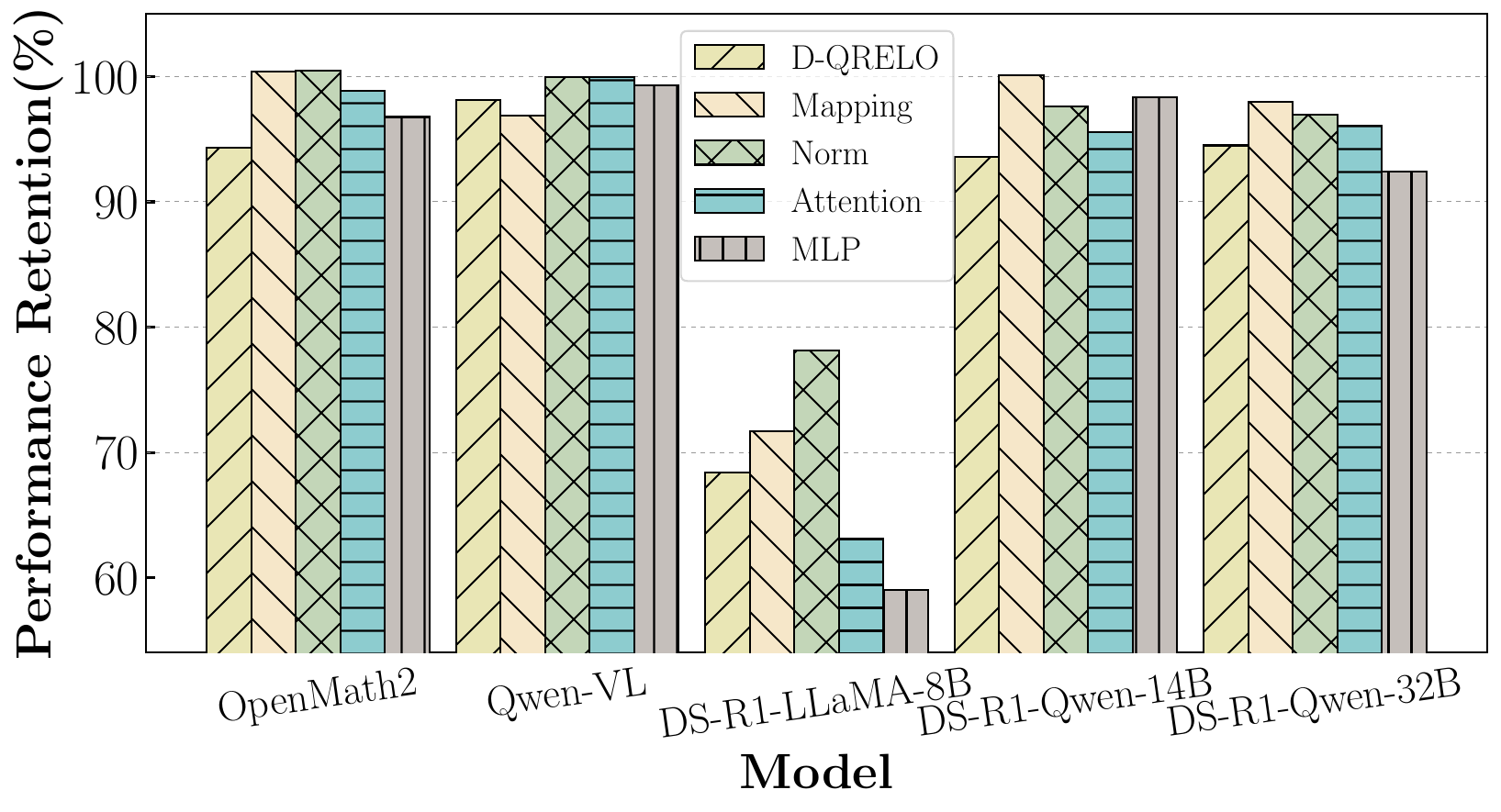}

		\caption{   Module sensitivity of delta compression. 
		}

		\label{fig:module}
	\end{figure}

\begin{figure}[t]
		\centering
		\includegraphics[width=0.99\columnwidth]{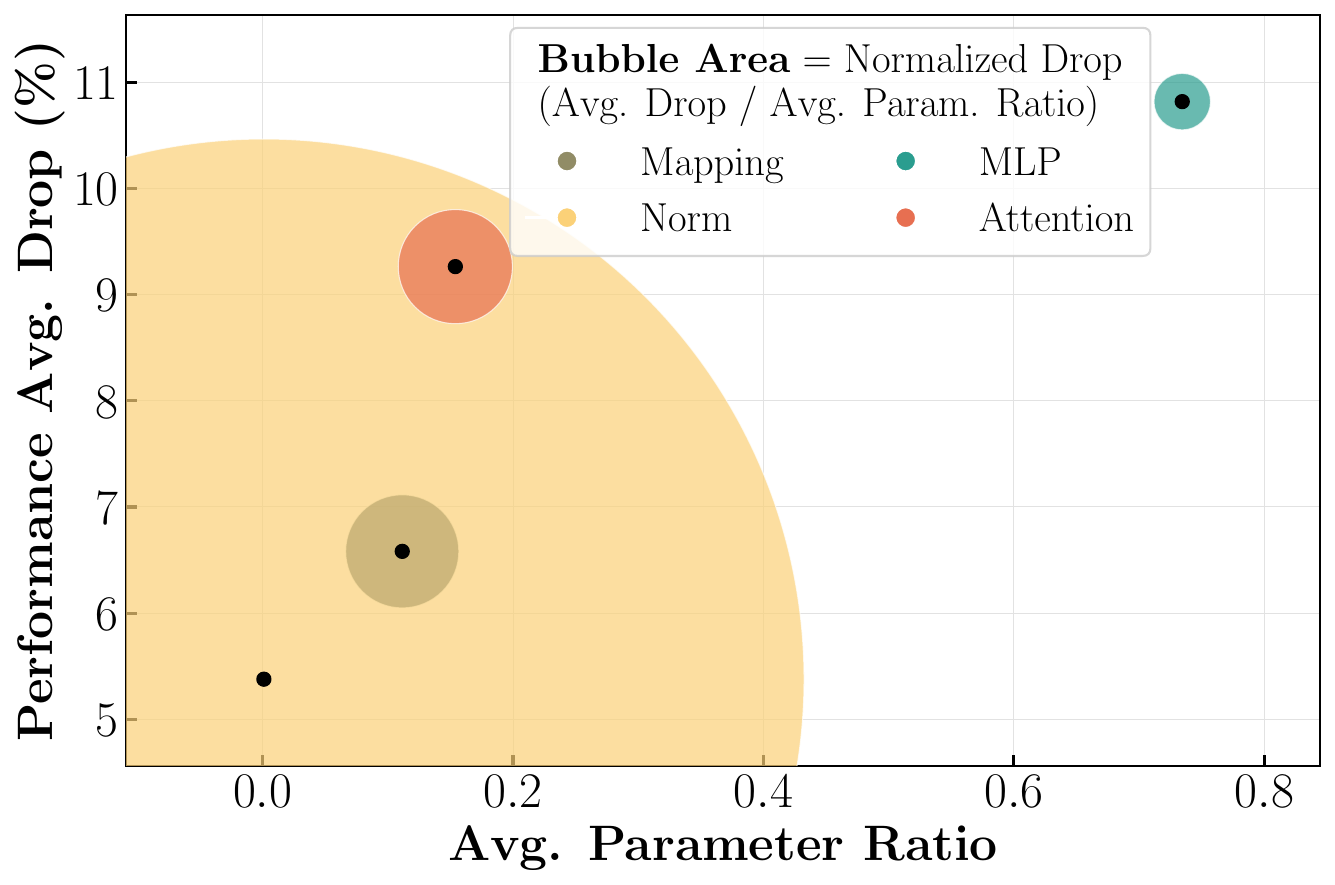}

		\caption{   Module importance in delta compression: Performance drop as a function of parameter ratio, where bubble size reflects normalized impact.
		}

		\label{fig:bubble}
	\end{figure}

To identify which modules of delta parameters are most sensitive to compression, we conduct an ablation study, applying \ourapproach selectively to different functional components: Mapping layers (token embedding and LM head), Normalization layers, Attention modules, and MLP modules. 
Figure~\ref{fig:module} shows the performance retention for each module type, while Figure~\ref{fig:bubble} summarizes the average performance drop relative to parameter ratio.

Firstly, module sensitivity to compression varies significantly across models.
DS-R1-LLaMA-8B stands out dramatically, exhibiting consistently lower performance retention, even when only a small fraction of its delta parameters (i.e., Normalization layers) are compressed. 
This is substantially lower than the near-perfect retention observed for Normalization layers in other models.

Secondly, MLP modules are the elephant in the room. They make up the vast majority of delta parameters and thus cause the largest absolute average performance drop. However, their moderate normalized drop indicates that while they are undeniably crucial, the performance decline is largely proportional to their size.

Thirdly, Normalization layers are the hidden vulnerability. Though they account for a negligible parameter ratio, they cause an exceptionally high normalized drop. 
This highlights that, despite their small size, Normalization layers are disproportionately sensitive to compression.

Finally, Mapping and Attention modules also exhibit a significant impact per parameter, with normalized drops second only to Normalization layers. This confirms their substantial sensitivity: despite being smaller than MLP modules, their impact per parameter compressed is pronounced.

\section{Conclusion}
In this paper, we investigate why previous delta compression methods suffer significant performance drops on powerful SFT models, especially those fine-tuned with large-scale datasets. 
We find that larger SFT data scale leads to increased magnitudes of delta parameters and their singular values, amplifying compression errors.
To tackle this, we propose \ourapproach, a training- and data-free delta compression method combining one-bit quantization with residual low-rank approximation. 
Experiments on several LLMs under this challenging setting show that \ourapproach outperforms key baselines. 
We also present a comprehensive empirical study of delta compression based on \ourapproach.

\section*{Acknowledgements}
This work was supported in part by National Natural Science Foundation of China (62476070), Shenzhen Science and Technology Program \seqsplit{(JCYJ20241202123503005, \, GXWD20231128103232001, \,ZDSYS20230626091203008,\, KQTD20240729102154066)}, Department of Science and Technology of Guangdong (2024A1515011540) and National Key R\&D Program of China (SQ2024YFE0200592).

\section*{Limitations}

Firstly, \ourapproach is constrained by a physical lower bound on its compression ratio because of its two-stage design. The total compression ratio $\rho$ is determined by the sum of the residual low-rank component $\rho_{1}$ and the one-bit quantization contribution $1/b$. For models using standard 16-bit precision (e.g., FP16 or BF16), the total ratio cannot be lower than $1/16$, which is approximately $6.25\%$, even if the residual component is negligible. 

Secondly, the relative advantages of \ourapproach tend to diminish when the SFT data scale is relatively small. Under these conditions, the resulting delta parameters are often sparse and exhibit smaller magnitudes. Hence, traditional and simpler compression methods, such as basic SVD or quantization, may already achieve high performance retention. Therefore, the added structural complexity of \ourapproach might not offer significant performance gains over more straightforward baselines.

\section*{Ethical Considerations}
Our research is conducted using publicly available and safe datasets and models. However, we explicitly acknowledge that the applicability of our \ourapproach and findings may be limited to datasets or domains similar to those studied. The performance of our approach on other specific datasets or domains remains uncertain, and there may be potential risks when applying it to privacy-sensitive or high-risk scenarios. Furthermore, the generalizability of our findings to real-world applications may require further exploration and testing. Therefore, caution is advised, and thorough verification is necessary to ensure the method generates accurate and reliable results in such contexts.

\bibliography{custom}
\newpage

\appendix
\clearpage
\input{table/appendix_models}

\section{Experiments Configuration}
\label{sec:lora_setting}
\input{table/models}

In our setup, LoRA parameters are matched with those of \ourapproach (1,040,187,392 vs. 1,003,782,656) to ensure fair comparison. Specifically, LoRA adopts a rank of 512 and an $\alpha$ of 512, applied to the Q, K, V, O projections as well as the input/output projections of FFN, without dropout or projection sharing/parallel strategies. For the optimization hyperparameters, LoRA fine-tuning uses a warmup ratio of 0.04 and a peak learning rate of 1e-4.
When evaluating OpenMath2 and Qwen2-VL-7B-Instruct, the temperature was set to 0, $top_p$ to 1.0, and maxtokens to 4096, with only one test performed.
However, for models in the DeepSeek-R1-distillate series, the temperature was set to 0.6, $top_p$ to 0.95, maxtokens to 20000, and four tests were conducted, and the results were averaged.
For Qwen3 models, the temperature was set to 0.7, $top_p$ to 0.8, $top_k$ to 20, and maxtokens to 20000, with one test conducted.

All experiments were conducted on a machine running Ubuntu 20.04, equipped with an NVIDIA A800-SXM4 80GB GPU and an Intel(R) Xeon(R) Platinum 8358P CPU @ 2.60 GHz. The system adopts CUDA 12.4, and the experiments were implemented using PyTorch 2.5.1.

\section{Evaluation Metrics}
All datasets are evaluated using accuracy as the primary metric, except for MME, which uses \text{Score} as the evaluation metric, two metrics are introduced for delta compression: \text{Drop} and \text{Performance Retention}.

The \text{Performance Retention} metric, representing the retained performance after compression, is defined as:
\begin{equation}
PR = \frac{\text{Compressed} - \text{Base}}{\text{SFT} - \text{Base}}
\end{equation}
where $PR$ stands for Performance Retention, while $\text{Compressed}$, $\text{Base}$, and $\text{SFT}$ represent the performance of the respective models.

The \text{Drop} metric, representing the performance loss rate, is defined as:
\begin{equation}
\text{Drop} = 1 - PR
\end{equation}

\section{Additional Results}

\label{sec:result}
Table~\ref{tab:appendix} reports the performance of DS-R1-LLaMA-8B, DS-R1-Qwen-14B, and DS-R1-Qwen-32B on MATH, AIME, GPQA, and HumanEval tasks. Consistent with our main findings, \ourapproach remains the top-performing method across all evaluated settings.

\end{document}

%% file: table/motivation.tex
\setlength{\tabcolsep}{0.4em}
\begin{table}[!t]
\begin{small}
\begin{center}
{
\begin{tabular}{l c c c}
\toprule
{\bf Data Size} & {\bf Avg($|\Delta|$) } & {\bf Avg($|\sigma|$)} &{\bf Avg($|H(\Delta)|$)}\\
\midrule
{\textsc{100K}}

& {\textsc{0.000492}}

& {\textsc{0.0181}}

& {\textsc{8.932}}
\\
{\textsc{300K}}

& {\textsc{0.001052}}

& {\textsc{0.0553}}
& {\textsc{9.348}}
\\
{\textsc{5M}}

& {\textsc{0.001489}}

& {\textsc{0.0967}}
& {\textsc{9.617}}
\\
{\textsc{14M}} 
 
& {\textsc{0.002147}}

& {\textsc{0.1266}}
& {\textsc{9.839}}

\\
\bottomrule
\end{tabular}
}
\end{center}
\end{small}
\caption{Impact of SFT data scale on delta parameters magnitude ($\Delta$), singular values ($\sigma$), entropy ($H(\Delta)$).}
\label{tab:motivation}
\vspace{-1.7mm}
\end{table}

%% file: table/main.tex
{
\setlength{\tabcolsep}{0.090em}

\def\arraystretch{1.41}
\begin{table*}[t]
\begin{center}
\begin{small}
\begin{tabular}{l|  cc| cc| cc| ccc| ccc| c}
\toprule
\multirow{2}{*}{\bf Method} 
& \multicolumn{2}{c|}{\text{OpenMath2}} 
& \multicolumn{2}{c|}{\text{Qwen2-VL}}
& \multicolumn{2}{c|}{\text{DS-R1-Qwen3-8B}}
& \multicolumn{3}{c|}{\text{Qwen3-4B-Instruct}}
& \multicolumn{3}{c|}{\text{Qwen3-30B-A3B-Instruct}}
& \multirow{2}{*}{\bf Drop}
\\
\cmidrule(lr){2-3} \cmidrule(lr){4-5} \cmidrule(lr){6-7} \cmidrule(lr){8-10} \cmidrule(lr){11-13}
 & {  GSM8K}
& { MATH} 
& { TextVQA} 
& {MME} 
& { AIME24} 
& { GPQA} 
& { MMLU} 
& { LCB} 
& { IFEval} 
& { MMLU} 
& { LCB} 
& { IFEval} 
& \\
\midrule
{Backbone} 

& {5.45} 
& {2.6}
& {-}
& {-}
& {7.78}
& {27.61}
& {18.25}
& {10.00}
& {34.57}
& {58.74}
& {16.79}
& {39.37}
& {-}
\\

{SFT} 

& {88.93} 
& {61.2}
& {84.51}
& {1676}
& {45.56}
& {61.28}
& {63.25}
& {30.40}
& {82.23}
& {74.13}
& {40.46}
& {83.18}

& {0\%}
\\
\midrule

{Random}
& {9.70} & {5.6} & {9.12} & {510} & {3.17} & {31.97} & {38.12} & {9.65} & {48.33} & {60.09} & {18.81} & {50.66}

& {86.12\%}
\\

{Magnitude}
& {45.56} & {25.2} & {17.37} & {871} & {18.67} & {41.49} & {52.75} & {13.62} & {71.89} & {67.16} & {22.67} & {74.21}

& {53.25\%}
\\

{Wanda}
& {50.34} & {27.6} & {15.79} & {784} & {20.13} & {40.34} & {50.91} & {15.24} & {73.15} & {66.31} & {24.83} & {76.08}

& {51.79\%}

\\

{SVD}
& {69.82} & {33.8} & {72.22} & {1451} & {21.50} & {47.31} & {54.05} & {16.08} & {75.42} & {69.45} & {23.04} & {76.30}

& {35.62\%}
\\
{BitDelta}

& {39.19} & {21.2} & {73.87} & {1525} & {8.13} & {39.28} & {47.98} & {10.69} & {67.76} & {64.44} & {14.37} & {71.92} 

& {56.13\%}
\\
{Delta-CoMe }

& \underline{80.50} & \underline{50.4} & \underline{75.69} & \underline{1569} & \underline{37.83} & \underline{57.06} & \underline{61.47} & \underline{23.13} & \textbf{80.51} & \underline{72.24} & \underline{32.72} & \underline{82.51} 

& \underline{14.00\%}
\\
{\textbf{\ourapproach}}

& \textbf{84.84} & \textbf{57.4} & \textbf{81.94} & \textbf{1664} & \textbf{42.33} & \textbf{59.65} & \textbf{62.71} & \textbf{24.91} & \underline{78.76} & \textbf{75.88} & \textbf{35.09} & \textbf{83.73}

& \textbf{6.17\%}
\\
\bottomrule
\end{tabular}
\end{small}
\end{center}

\caption{The performance of different delta-compression methods on five LLMs. LCB denotes LiveCodeBench. Optimal results are bolded, sub-optimal ones are underlined. Drop means the percentage of performance degradation.
}
\label{tab:main}
\end{table*}
}

%% file: table/ab1.tex
{
\setlength{\tabcolsep}{0.142em}
\def\arraystretch{1.23}
\begin{table*}[t]
\begin{center}
\begin{small}
\begin{tabular}{l|  cc| cc| cc| ccc| ccc| c}
\toprule
\multirow{2}{*}{\bf Method} 
& \multicolumn{2}{c|}{\text{OpenMath2}} 
& \multicolumn{2}{c|}{\text{Qwen2-VL}}
& \multicolumn{2}{c|}{\text{DS-R1-Qwen3-8B}}
& \multicolumn{3}{c|}{\text{Qwen3-4B-Instruct}}
& \multicolumn{3}{c|}{\text{Qwen3-30B-A3B-Instruct}}
& \multirow{2}{*}{\bf Drop}
\\
\cmidrule(lr){2-3} \cmidrule(lr){4-5} \cmidrule(lr){6-7} \cmidrule(lr){8-10} \cmidrule(lr){11-13}
 & {  GSM8K}
& { MATH} 
& { TextVQA} 
& {MME} 
& { AIME24} 
& { GPQA} 
& { MMLU} 
& { LCB} 
& { IFEval} 
& { MMLU} 
& { LCB} 
& { IFEval} 
& \\
\midrule
{SVD} 

& {43.97} 
& {26.8}
& {57.19}
& {1280}
& {13.55}
& {37.27}
& {43.71}
& {11.06}
& {58.11}
& {64.80}
& {19.34}
& {65.08}

& {58.71\%}

\\

\small{+ One-bit}
& \underline{80.64} & \underline{50.8} & \underline{80.76} & \underline{1652} & \textbf{42.71}
& \underline{58.53}
& \underline{62.49}
& \underline{24.37}
& \textbf{78.85}
& \underline{74.52}
& \textbf{35.26}
& \underline{82.92}

& \underline{9.01\%}
\\

\small{+ SVD}
& {69.82} & {33.8} & {72.22} & {1451} & {21.50} & {47.31} & {54.05} & {16.08} & {75.42} & {69.45} & {23.04} & {76.30}

& {35.62\%}

\\
\midrule

{One-bit} 

& {39.19} & {21.2} & {73.87} & {1525} & {8.13} & {39.28} & {47.98} & {10.69} & {67.76} & {64.44} & {14.37} & {71.92} 

& {56.13\%}
\\

{+ SVD}
& \textbf{84.84} & \textbf{57.4} & \textbf{81.94} & \textbf{1664} & \underline{42.33} & \textbf{59.65} & \textbf{62.71} & \textbf{24.91} & \underline{78.76} & \textbf{75.88} & \underline{35.09} & \textbf{83.73}

& \textbf{6.17\%}
\\

\bottomrule
\end{tabular}
\end{small}
\end{center}

\caption{ Ablation study of \ourapproach's two-stage strategy. Each method first applies SVD or one-bit quantization to the delta parameters ($\rho$ = 1/16). The $+$ represent compression of the residuals by SVD or one-bit quantization. 
}
\label{tab:ab1}

\end{table*}
}

%% file: table/memory.tex
\begin{table}[t]
\centering 
\small 
\setlength{\tabcolsep}{4pt} 
\begin{tabular}{l c c} 
\toprule
\textbf{\# Models} & \textbf{Memory (GB)} & \textbf{Latency (ms)} \\
& ({$\times$} \ / \ \checkmark) \ourapproach & ({$\times$} \ / \ \checkmark) \ourapproach \\ 
\midrule
2              & 31.78 / 20.91               & 63 / 45   \\
4              & 63.12 / 26.88               & 129 / 52   \\
8              & OOM / 38.94  & - / 67 \\
16             & OOM / 62.87  & - / 91 \\ 
\bottomrule
\end{tabular}
\caption{GPU memory cost and inference latency.}
\label{tab:memory_latency}
\end{table}

%% file: table/lora.tex
\begin{table}[t]
\begin{small}
\begin{center}
{
\begin{tabular}{l c c c}
\toprule
\multirow{2}{*}{\bf Method} 
& \multicolumn{2}{c}{{OpenMath2-LLaMA3.1-8B}}  
&\multirow{2}{*}{\bf Avg.}\\
\cmidrule(lr){2-3} 
& {\bf  GSM8K} & {\bf  MATH} \\
\midrule

{{Backbone}} 
& {\textsc{5.45}} 
& {\textsc{2.6}}
& {\textsc{4.03}}
\\
{{SFT}}
& {\textsc{88.93}}
& {\textsc{61.2}}
& {\textsc{75.07}}
\\
\midrule
{{LoRA}}
& {\textsc{74.37}}
& {\textsc{49.8}}
& {\textsc{62.09}}
\\
{{SVD}}
& {\textsc{69.82}}
& {\textsc{33.8}}
& {\textsc{51.81}}
\\
{\ourapproach}
& {\textsc{84.84}}
& {\textsc{57.4}}
& {\textsc{71.12}}
\\

\bottomrule
\end{tabular}
}
\end{center}
\end{small}
\caption{LoRA vs. delta-compression.}
\label{tab:lora}
\end{table}

%% file: table/appendix_models.tex
\begin{strip}
\centering
\setlength{\tabcolsep}{0.091em}
\def\arraystretch{1.41}
\begin{small}
\begin{tabular}{l|ccc| cccc| ccc}
\toprule
\multirow{2}{*}{\bf Method} 
& \multicolumn{3}{c|}{\text{DS-R1-LLaMA-8B}}
& \multicolumn{4}{c|}{\text{DS-R1-Qwen-14B}}
& \multicolumn{3}{c}{\text{DS-R1-Qwen-32B}}

\\
\cmidrule(lr){2-4} \cmidrule(lr){5-8} \cmidrule(lr){9-11}

& { MATH} 
& { AIME24} 
& { GPQA} 
& { MATH} 
& { AIME24} 
& { GPQA} 
& { HumanEval}
& { MATH} 
& { AIME24} 
& { GPQA} 
\\
\midrule
{Backbone}

& {2.60}
& {0}
& {20.13}
& {64.15}
& {6.67}
& {32.32}
& {72.6}
& {62.40}
& {8.33}
& {36.36}
\\

{SFT}

& {89.55}
& {44.17}
& {45.20}
& {92.25}
& {61.67}
& {56.94}
& {77.4}
& {92.65}
& {67.50}
& {63.51}

\\
\midrule

{Random}
 & {7.65} & {0} & {22.22} & {58.45} & {6.67} & {30.39} & {69.5} & {57.30} & {8.17} & {34.93}

\\

{Magnitude}
 & {18.40} & {1.33} & \underline{31.19} & {76.35} & {17.33} & {37.82} & {75.2} & {78.65} & {20.83} & {42.74}

\\

{Wanda}
& {24.70} & {2.17} & {30.72} & {78.20} & {17.67} & {39.04} & \underline{75.8} & {78.40} & {20.83} & {44.52}

\\

{SVD}
 & {39.20} & {4.67} & {30.32} & {53.90} & {15.50} & {44.34} & {69.5} & {60.30} & {17.67} & {50.43}

\\
{BitDelta}
 & {11.80} & {0.83} & {30.56} & {20.95} & {0} & {35.35} & {73.8} & {22.75} & {0} & {39.43}

\\
{Delta-CoMe }

 & \underline{45.40} & \underline{5.83} & {30.18} & \underline{87.70} & \underline{47.00} & \underline{51.92} & {66.5} & \underline{88.10} & \underline{53.33} & \underline{58.46}

\\
{\textbf{\ourapproach} \textbf{(ours)}}

 & \textbf{79.05} & \textbf{23.83} & \textbf{35.98} & \textbf{91.40} & \textbf{57.50} & \textbf{53.27} & \textbf{77.4} & \textbf{92.45} & \textbf{62.50} & \textbf{61.47}

\\
\bottomrule
\end{tabular}
\end{small}
\captionof{table}{The performance of different delta-compression methods on three powerful SFT LLMs. Optimal results are in bold, while sub-optimal results are underlined.}
\label{tab:appendix}
\end{strip}

%% file: table/models.tex
\begin{table}[ht]
\centering
\small
\renewcommand{\arraystretch}{1.15}  
\begin{tabular}{@{}lc@{}}
\toprule
\textbf{Model}  & \textbf{Backbone}  \\
\midrule
OpenMath2-LLaMA3.1-8B        & LLaMA3.1-8B        \\
Qwen2-VL-7B-Instruct         & Qwen2-7B           \\
DeepSeek-R1-Distill-LLaMA3-8B & LLaMA3.1-8B        \\
DeepSeek-R1-Distill-Qwen-14B & Qwen2.5-14B        \\
DeepSeek-R1-Distill-Qwen-32B & Qwen2.5-32B        \\
DeepSeek-R1-0528-Qwen3-8B    & Qwen3-8B-Base      \\
Qwen3-4B-Instruct-2507       & Qwen3-4B-Base      \\
Qwen3-30B-A3B-Instruct-2507  & Qwen3-30B-A3B-Base \\
\bottomrule
\end{tabular}
\caption{Backbone of each evaluated model.}
\label{tab:models}
\end{table}